\documentclass[journal]{IEEEtran}

\usepackage{color}

\ifCLASSINFOpdf
\else
\fi

\usepackage{graphicx}

\usepackage{graphics}
\begin{document}
\title{SCAttNet: Semantic Segmentation Network with Spatial and Channel Attention Mechanism for High-Resolution Remote Sensing Images}
\vspace{-0.4cm}

\author{\author{Haifeng Li,~\IEEEmembership{Member,~IEEE,}
        Kaijian Qiu, Li Chen, Xiaoming Mei, Liang Hong,
        and~Chao Tao* % <-this % stops a space
\thanks{Haifeng Li, Kaijian Qiu, Li Chen, Xiaoming Mei,
        and Chao Tao are with the School of Geosciences and Info-Physics, Central South University, Changsha 410083 China. Liang Hong is with the College of Tourism and Geographic Science, Yunnan Normal University, Kunming, China. Corresponding author: Chao Tao, e-mail: kingtaohao@csu.edu.cn.}% <-this % stops a space
% <-this % stops a space
}% <-this % stops a space
\thanks{Manuscript received July 19, 2019.}}


\markboth{IEEE of \LaTeX\ Class Files,~Vol.~14, No.~8, August~2019}%
{Shell \MakeLowercase{\textit{et al.}}: Bare Demo of IEEEtran.cls for IEEE Journals}



\maketitle


\begin{abstract}
High-resolution remote sensing images (HRRSIs) contain substantial ground object information, such as texture, shape, and spatial location. Semantic segmentation, which is an important task for element extraction, has been widely used in processing mass HRRSIs. However, HRRSIs often exhibit large intraclass variance and small interclass variance due to the diversity and complexity of ground objects, thereby bringing great challenges to a semantic segmentation task. In this paper, we propose a new end-to-end semantic segmentation network, which integrates  lightweight spatial and channel  attention modules that can refine features adaptively. We compare our method with several classic methods on the ISPRS Vaihingen and Potsdam datasets. Experimental results show that our method can achieve better semantic segmentation results. The source codes are available at https://github.com/lehaifeng/SCAttNet.
\end{abstract}


% \fi
\IEEEpeerreviewmaketitle

\begin{figure*}[t]
    \vspace{-0.4cm}
	\centering
	\includegraphics[width=5.5in,height=1.9in]{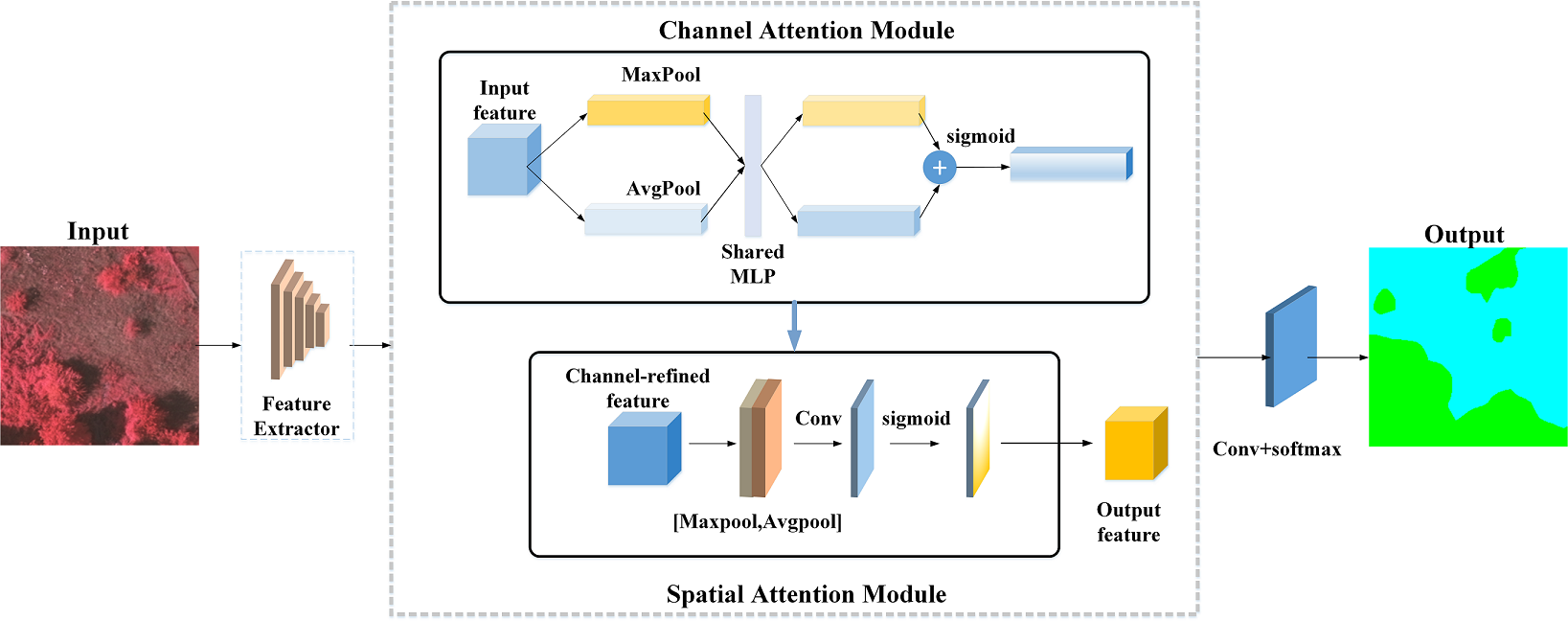}
	\caption{Overview of our proposed SCAttNet network}
	\label{fig1}
	\vspace{-0.4cm}
\end{figure*}

\section{Introduction}
\IEEEPARstart{S}{emantic} segmentation of remote sensing images is a fundamental task that classifies each pixel in an image into a specified category. It plays an important role in many fields such as change detection, element extraction, and military target recognition. \par

Image semantic segmentation methods can be divided into two categories: traditional methods and deep learning based ones. Traditional methods use the color, texture, shape, and spatial position relationships of an object to extract features and then use clustering, classification, and threshold algorithms to segment an image \cite{huang2011information}\cite{yang2012layered}. However, these methods depend heavily on artificial design features and show some bottlenecks. Recently, deep learning based methods have been regarded as a promising approach to solve image semantic segmentation problems \cite{ronneberger2015u}\cite{long2015fully}\cite{badrinarayanan2017segnet}. For example, methods based on fully convolutional network (FCN) \cite{long2015fully} have achieved state-of-the-art segmentation results on many natural image datasets, such as PASCAL VOC \cite{everingham2010pascal} and Cityscapes \cite{cordts2016cityscapes}.\par

However, remote sensing images are different from natural images and often viewed from a high-altitude angle. Thus, the range of imaging is wide, and the background is complex and diverse. Especially in HRRSIs, the difference of ground objects becomes further notable. To segment HRRSIs effectively, many advanced methods have been proposed. For example, weighted focal loss \cite{dong2019denseu} and multi-class loss combined by Dice loss and Binary Cross Entropy loss \cite{zhu2019efficient} have been proposed to solve the class imbalance problem in remote sensing images. Multi-modal datas such as digital surface models (DSMs) \cite{chen2019effective}\cite{audebert2016semantic} have been used to improve semantic segmentation performance. In addition, spatial relation module \cite{mou2019spatial} and spatial information inference structure \cite{tao2019spatial} have been designed to model more effective contextual spatial relations. These methods have achieved satisfactory results for remote sensing images segmentation.\par
 
Recently, attention mechanisms have been successfully applied to semantic segmentation. These methods can be divided into two categories. One uses attention mechanism to select meaningful features at channel dimension. For example, PAN\cite{li2018pyramid} uses a global attention module as a context prior to select precise features channel-wisely. Attention U-net \cite{oktay2018attention} uses a channel attention module to control the fusion of high-level and low-level features at channel dimension. However, these method do not consider to enhance the feature representation at spatial dimension. Another one is called the self attention mechanism, which calculates the feature representation in each position by weighted sum the features of all other positions \cite{fu2019dual}\cite{huang2019ccnet}. Thus, it can model the long-range context information of semantic segmentation task. For example, DANet \cite{fu2019dual} uses two self-attention modules to model long-range context information at spatial and channel  dimension respectively. However, these methods are complexity in the size of model, thus yield inefficient computation which will be great challenge to process massive remote sensing images. In our work, two lightweight attention modules \cite{woo2018cbam} which contains spatial attention and channel attention are adopted for the semantic segmentation of HRRSIs. The spatial attention module models the spatial features of HRRSIs, and the channel attention module captures what to enhance. Integrating these two attention modules can effectively improve the accuracy of semantic segmentation. The contributions of this study primarily include the following three points:\par
(1) We propose a new semantic segmentation Network with Spatial and Channel Attention (SCAttNet) to improve the semantic segmentation accuracy of HRRSIs.\par
(2) We visualize the representations learned by spatial and channel attention modules to explain why does our method work.\par
(3) Experiments on Vaihingen and Potsdam datasets demonstrate competitive results by learning representations for HRRSIs via spatial and channel attention modules, and show remarkable performance improvements on small objects.\par

\section{Method}

\subsection{Overview of SCAttNet}
The proposed semantic segmentation network is shown in Fig. \ref{fig1}. It consists of two parts: a backbone for feature extraction and an attention module. The attention module is composed of a channel attention and a spatial attention in cascade. For an input remote sensing image, we first use the backbone network for feature extraction. Then we feed the extracted feature map into the channel attention module to refine the features in channels. Afterward, we feed the refined channel feature map into the spatial attention module for refinement at spatial axis. Lastly, we can obtain the  semantic segmentation results via convolution and SoftMax operations. Specific network details and design ideas are shown in the following sections.

\subsection{Backbone for feature extraction}
 In this study, we use two representative backbones for feature extraction: SegNet and ResNet50 \cite{he2016identity}. On this basis, we proposed two networks: SCAttNet V1 with SegNet as backbone and SCAttNet V2  with ResNet50 as backbone. For SegNet network, it has been widely used as the baseline model of remote sensing images semantic segmentation and achieved sound semantic segmentation results. For example, Audebert et al. \cite{audebert2016semantic} based on SegNet network, combined with multi-modal data, achieved state-of-the-art semantic segmentation results on the ISPRS Vaihingen dataset. And ResNet50 is also a common backbone for  semantic  segmentation tasks because it can build a deep-layer model with a wide receptive field. We use eight times downsampling for ResNet50 in our work. Unlike the SegNet and ResNet50 networks, we do not directly use the features of the last layer for semantic inference, instead, we feed the feature map of the last layer into the attention module for feature refinement and then make semantic inference, which is conducive to learning better feature expression.\par

\subsection{Attention Module}
\textbf{Channel Attention}: Given a high-resolution remote sensing image, it will produce a multichannel feature map $F\in R^{C\times H \times W}$ (where C, H, and W denote the number of channels, the height, the width of the feature map, respectively) after passing through several convolutional layers. The information expressed in the feature map of each channel is different. Channel attention aims to use the relationships between each channel of the feature map to learn a 1D weight $W_{c}\in R^{C\times 1 \times 1}$, and then multiply it to the corresponding channel. In this manner, it can pay more  attention to the meaningful semantic information for the current task. To learn effective weight representation, we first aggregate spatial dimension information through global average pooling and global max pooling to generate two feature descriptors for each channel. Then we feed the two feature descriptors into a shared multilayer perceptron with one hidden layer (where the number of the hidden layer units is C/8) to generate more representative feature vectors. Afterward, we merge the output feature vectors through an element-wise summation operation. Finally, using a sigmoid function, we can obtain the final channel attention map. The flow chart is illustrated in the channel attention module of Fig. 1. The formula for calculating channel attention is shown in Formula 1:\par

\begin{equation}\footnotesize
\vspace{-0.1cm}
    W_{c}(F)=sigmoid(MLP(AvgPool(F))\\
    +MLP(Maxpool(F)))
\end{equation}

\textbf{Spatial Attention}: For the spatial attention, it focuses on where are valuable for current tasks. In HRRSIs, the ground objects are exhibited in various sizes and the distribution is complicated. Therefore, using spatial attention is useful for aggregating spatial information, especially for small ground objects. Spatial attention utilizes the relationships between different spatial positions to learn a 2D spatial weight map $W_{s}$ and then multiplies it to the corresponding spatial position to learn more representative features. To learn the spatial weight relationships effectively, we first generate two feature descriptors for each spatial position through global average pooling and global max pooling operations. Then we concatenate two feature descriptors together and generate a spatial attention map through a $7\times7$ convolution operation. Lastly, we use a sigmoid function to scale the spatial attention map to $0\sim1$. The flow chart is illustrated in the spatial attention module of Fig. 1. The spatial attention calculation formula is shown in Formula 2.\par

\begin{equation}\small
%{\setlength\abovedisplayskip{0pt}
 %\setlength\belowdisplayskip{0pt}
    W_{s}(F)=sigmoid(f^{7\times7}([Avgpool(F);Maxpool(F)]))
%}
\end{equation}

where $f^{7\times7}$ represents a convolution operation with $7\times7$ kernel size.\par

 We can easily compute that the number of parameters of the attention modules are so small that they can be neglected for both our SCAttNet V1 and SCAttNet V2, which have tens of millions of parameters. Specifically, the number of parameters of the channel attention module is $C \times C/4$(C is the number of channels) and the and the number of parameters of the spatial attention module is 98. In addition, we only add the attention module at the end layer of the backbone network; thus, it seldom brings computation complexity to our proposed network.\par

In this study, we follow the method of Woo \cite{woo2018cbam}to integrate the two attention modules. First, we use channel attention to select significant feature maps in each layer, then use spatial attention to choose considerable neuron's activity in each feature maps, which is conducive to extracting more valuable feature expression.

\begin{table*}[tbp]
    \vspace{-0.4cm}
	\centering  % 表居中
	\caption{ Semantic segmentation results of ISPRS Vaihingen dataset. The accuarcy of each category is presented in the IoU/F1-score form.}
	
	\begin{tabular}{ccccccccc}  % {lccc} 表示各列元素对齐方式，left-l,right-r,center-c
		\hline
	Model	&Imp. Surfaces	&Building	&Low  Veg.	&Tree	&Car	&MIoU (\%)	&AF (\%)	&OA(\%) \\ \hline
	
	FCN-32s \cite{long2015fully}	&68.31/81.17	&70.59/82.76	&57.22/72.79	&60.95/75.74	&8.16/15.09	&53.04	&65.51	&78.01 \\
	
	FCN-8s \cite{long2015fully}	&71.32/83.26	&71.64/83.48	&63.67/77.80	&65.78/79.36	&44.73/61.81	&63.43	&77.14	&80.99 \\
	
	U-net \cite{ronneberger2015u}	&79.48/88.57	&81.91/90.06	&65.03/78.81	&66.78/80.08	&52.36/68.74	&69.11	&81.25	&84.88 \\
	
	SegNet \cite{badrinarayanan2017segnet}	&75.73/86.19	&79.09/88.32	&62.83/77.17	&65.42/79.09	&37.22/54.24	&64.06	&77.00	&82.92  \\
	
	SegNet + cha. att	&76.74/86.84	&80.46/89.17	&63.31/77.53	&65.13/78.89	&41.67/58.83	&65.46	&78.25	&83.48  \\
	
	SegNet + spa. att	&74.46/85.36	&77.01/87.01	&63.37/77.58	&65.08/78.85	&40.30/57.45	&64.05	&77.25	&82.38   \\
	
	SCAttNet V1(ours)	&77.75/87.36	&81.05/89.54	&63.00/77.30	&65.51/79.16	&47.71/64.60	&66.96	&79.59	&83.79   \\
	
	ResNet50 & 78.46/87.93 &80.92/89.46&65.99/79.51&	65.84/79.41&53.68/69.86&68.99 & 81.23 & 84.57		
	 \\
	 
   CBAM \cite{woo2018cbam} &78.36/87.86 &80.56/89.23	&\textbf{66.97}/\textbf{80.22} &\textbf{68.91}/\textbf{81.60} &52.52/68.87	&69.46	&81.56	&84.96 \\
     
	RefineNet \cite{lin2017refinenet}& 77.42/87.27 &76.83/86.90 &64.40/78.34 &66.16/79.63 &\textbf{61.15}/\textbf{75.89} &69.19 &81.61 &83.36 \\
	
	DeepLabv3+ \cite{chen2018encoder}&79.43/88.54 &81.73/89.95 &\underline{66.88/80.15} &66.73/80.05 &52.57/68.91 &69.47 &81.52 &85.15\\
	
	G-FRNet \cite{amirul2017gated} &\underline{80.09/88.94} &\textbf{82.60}/\textbf{90.47} &66.85/80.13 &\underline{67.31/80.46} &\underline{56.99/72.60} &\textbf{70.77} &\textbf{82.52} &\textbf{85.52} \\

	SCAttNet V2(ours)& \textbf{80.40}/\textbf{89.13}&	\underline{82.32/90.30} &	66.73/80.04&	67.09/80.31&	54.44/70.50&	\underline{70.20}&	\underline{82.06}&	\underline{85.47} \\
	\hline
	\label{tabel1}
	\end{tabular}
	\vspace{-0.5cm}
\end{table*}

\begin{table*}[tbp]
	\centering  % 表居中
	\caption{Semantic segmentation results of ISPRS Potsdam dataset. The accuarcy of each category is presented in the IoU/F1-score form.}
	\begin{tabular}{cccccccccc}  % {lccc} 表示各列元素对齐方式，left-l,right-r,center-c
		\hline
	Model&Imp. Surfaces&Building&Low Veg.&Tree&Car&Clutter/background	&MIoU(\%)&AF (\%)&OA(\%)\\ \hline
	
	FCN-32s \cite{long2015fully}	&64.28/78.26	&71.78/83.57	&55.21/71.15	&49.39/66.12	&58.83/74.08	&9.58/17.49	&51.51	&65.11	&75.94\\
	
	FCN-8s \cite{long2015fully}	&66.81/80.10	&74.17/85.17	&57.19/72.77	&51.57/68.05	&69.02/81.67	&13.11/23.18	&55.31	&68.49	&77.82\\
	
	U-net \cite{ronneberger2015u}	&65.29/79.00	&73.64/84.82	&65.11/78.87	&59.88/74.91	&77.72/87.47	&14.97/26.04	&59.44	&71.85	&79.84\\
	
	SegNet \cite{badrinarayanan2017segnet}	&68.17/81.07	&76.07/86.41	&63.91/77.98	&58.54/73.85	&75.01/85.72	&13.41/23.65	&59.18	&71.45	&80.27\\
	
	SCAttNet V1(ours)	& 69.51/82.01	&77.40/87.26	&66.72/80.03	&62.50/76.92	&76.20/86.49	&15.22/26.42	&61.26	&73.19	&81.62 \\
	
    ResNet50 &\underline{81.25/89.65} &\underline{87.39/93.27} &71.13/83.13 &\underline{65.41/79.09} &\textbf{80.58}/\textbf{89.24} &17.39/29.62 &\underline{67.19} &\underline{77.34} &\underline{86.94} \\
	
	CBAM \cite{woo2018cbam} &79.52/88.60 &87.03/93.07	&68.96/81.63 &65.11/78.87 &79.86/88.81	&\underline{18.84/31.71}	&66.56	&77.11	&86.24 \\
	
	RefineNet\cite{lin2017refinenet} &77.91/87.58 &79.37/88.50 &69.36/81.91 &65.38/79.07 &78.41/87.90 &16.27/27.98 &64.45 &75.49 &84.38 \\
	
	DeepLabv3+ \cite{chen2018encoder}& 80.70/89.31 &86.59/92.81 &\underline{71.48/83.37} &64.47/78.40 &78.96/88.24 
	&18.74/31.56 &66.82 &77.28 &86.76 \\
	
	G-FRNet \cite{amirul2017gated}& 80.50/89.20 &86.38/92.69 &70.71/82.84 &65.29/79.00 &75.88/86.28 &18.05/30.58 &66.14 &76.77 &86.84 \\
	
	SCAttNet V2(ours) &\textbf{81.83}/\textbf{90.04} &\textbf{88.76}/\textbf{94.05} &\textbf{72.49}/\textbf{84.05} 
	&\textbf{66.33}/\textbf{79.75} &\underline{80.28/89.06} &\textbf{20.18}/\textbf{33.58} &\textbf{68.31} &\textbf{78.42} &\textbf{87.97} \\
	
	\hline
	\label{table2}
	\end{tabular}
	\vspace{-0.5cm}
\end{table*}

\section{Experiment}
In this section, we  evaluate the performance of our network on  the ISPRS Vaihingen and ISPRS Potsdam datasets \footnote{ ISPRS 2d semantic labeling dataset.http://www2.isprs.org/ \par commissions/comm3/wg4/semantic-labeling.html}.

\subsection{ Datasets and evaluation metrics}
The ISPRS Vaihingen dataset contains 33 orthophoto maps and related DSMs. Sixteen of them are labeled. The average size of the images is $2494\times2064$ and the resolution is 9 cm. Each image contains three bands: near infrared, red  and green bands. Moreover, it includes six categories: impervious surfaces, buildings, low vegetation, trees, cars and clutter/background. The ISPRS Potsdam dataset contains 38 orthophoto maps and related normalized DSMs. Twenty-four of them are labeled. The size of each image is $6000\times6000$ and the resolution is 5 cm. Each image contains four bands: near infrared, red, green  and blue bands. It has the similar categories as the ISPRS Vaihingen dataset.\par
To  evaluate our proposed model, we use three evaluation metrics including mean inter-section over union (MIoU), average F1-score (AF) and overall accuracy (OA) to evaluate semantic segmentation performance. 

\subsection{Implementation details}
 Considering that many remote sensing datasets in practice do not have DSMs, we do not use such datasets in this experiment for wide application value. For the ISPRS Vaihingen dataset, we divide the labeled dataset into two parts, in which (ID 30, 32, 34, 37) are for evaluating the performance of the network, and the remaining 12 images are for training. The number of Vaihingen dataset is small; thus, to prevent overfitting, we first crop the training dataset randomly into a $256\times256$ size and then expand the data through rotation and translation operations, and finally obtain 12,000 patches for training. The Potsdam dataset is also divided  into two parts, in which (ID 2\_12, 3\_12, 4\_12, 5\_12, 6\_12, 7\_12) are for testing, and the remaining 18 images are for training. Then, we crop it randomly and obtain
 27,000 $256\times256$ patches for training.
 
We trained all the model from the scratch without bells and whistles. The models including FCN-32s, FCN-8s, U-net, SegNet and G-FRNet all use VGG-16 as backbone. As for RefineNet and DeepLabv3+, we adopt the ResNet50 as backbone with 32 times downsampling for RefineNet and eight times downsampling for DeepLabv3+ as the original paper. We also adopt ResNet50 as backbone with eight times downsampling for CBAM. To train the proposed SCAttNet V1,  we set the learning rate of SCAttNet V1 on the Vaihingen and Potsdam datasets to 1e-3 and 1e-4, respectively. To train SCAttNet V2,  we set the learning rate of SCAttNet V2 on both datasets  to 1e-3. The proposed models all adopt Adam as the optimizer and cross-entropy as the loss function. And the training epochs is set to 50. Considering limited computing resources, the batch size is set to 16. To test all the above models, we only use a sliding window without overlap to crop the images and then stitch together. We conduct all our experiments in Tensorflow platform with a NVIDIA 1080Ti GPU.
 
 \subsection{Results of Vaihingen dataset}
Table \ref{tabel1} reports the semantic segmentation of the ISPRS Vaihingen dataset. We adopt the practice of \cite{liu2017hourglass}\cite{guo2018pixel} and do not report the accuracy of the clutter/background class because the Vaihingen dataset has less clutter/background. From the experimental results in Table 1, the MIoU/AF/OA increased by 1.4\%/1.25\%/0.56\%  with the channel attention module, compared with original SegNet. This increase shows the effectiveness of the channel attention module to capture meaningful semantic information; With the spatial attention module, the IoU/F1-score for small objects such as cars increased by 3.08\%/3.21\%. This increase shows that the spatial attention module can aggregate more location information even if the small objects in original images are 32 times downsampling in SegNet. Though the accuracy of impervious surfaces, buildings, and trees classes  slightly decreased,  we can still increase the MIoU/AF  by 2.9\%/2.59\% compared with original SegNet after inputting the channel-refined features to the spatial attention module. Moreover, in SCAttNet V2, the MIoU/AF/OA increased by 1.21\%/0.83\%/0.90\% compared with ResNet50 which as a baseline model. In addition, we find that in FCN-32s, the car category shows a low accuracy, but the car category in FCN-8s has improved much compared with FCN-32s, which shows the importance of  low-level features for segmenting small objects.
\par

To compare the semantic segmentation results, we visualize the semantic segmentation results of ID32. The visualization results are shown in the first row of Fig. \ref{fig2}. We can see that our SCAttNet V1 achieves more coherent results compared with the original SegNet model. In addition, our model infers more accurately in small object areas such as cars in the right middle of the image. We also test the accuracy of the ID32 area. The MIoU/AF/OA of SegNet is 60.57\%/74.01\%/82.78\%, whereas that of our SCAttNet V1 is 64.92\%/77.74\%/85.49\%. On this basis, our network can improve semantic segmentation accuracy by incorporating the attention module effectively.

\subsection{Results of Potsdam dataset}
Table \ref{table2} reports the semantic segmentation results of the ISPRS Potsdam dataset. From the semantic segmentation results, the proposed SCAttNet V2 model is superior to the  comparative models. Moreover, it achieves 1.12\%/1.08\%/1.03\% higher in MIoU/AF/OA metrics compared with ResNet50 which as a baseline model. In SCAttNet V1, The MIoU/AF/OA has increased by 2.08\%, 1.74\% and 1.35\%, respectively compared with SegNet which as a baseline model. Thus, we further confirm the effectiveness of the attention module.\par

\begin{figure}[t]
	\centering
	\includegraphics[width=3.5in]{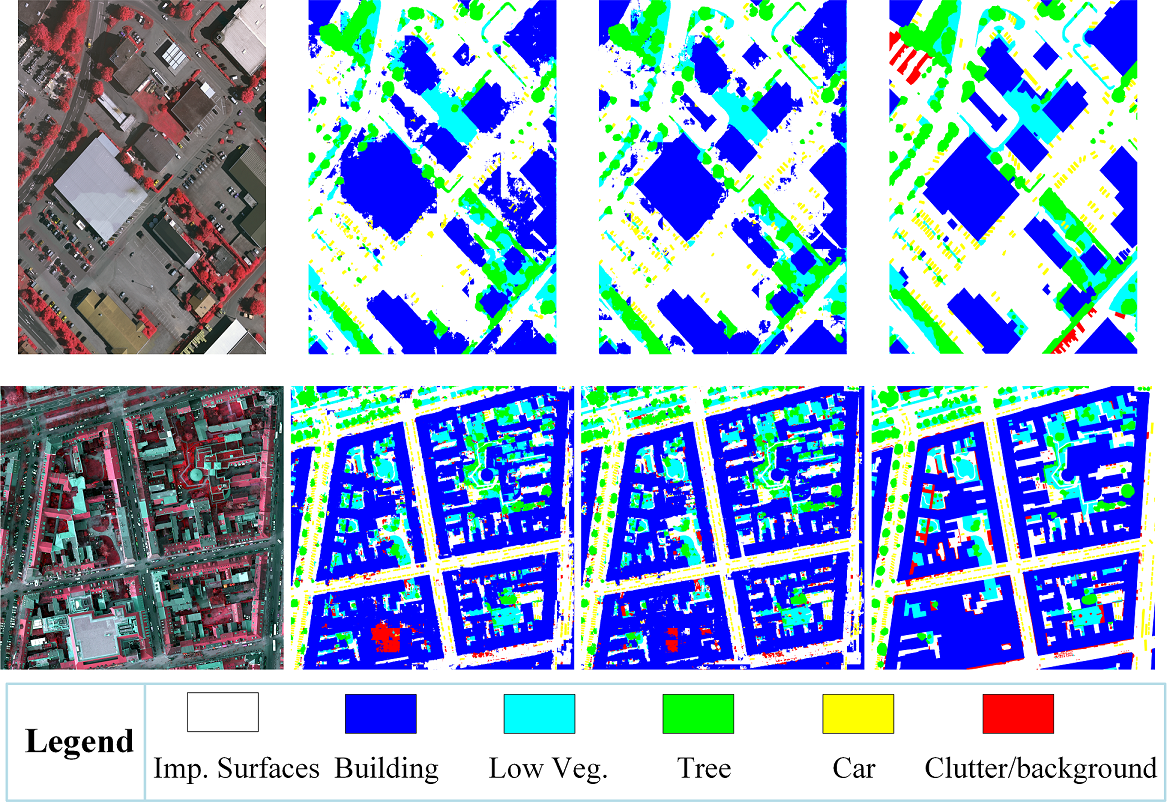}
	\caption{Visualization results of the Vaihingen and Potsdam datasets. The first row presents the visualization results of the ISPRS Vaihingen dataset, from left to right: original image, SegNet output results, SCAttNet V1 output results and ground truth. The second row presents the visualization results of the ISPRS Potsdam dataset, from left to right: original image, SegNet output results, SCAttNet V1 output results and ground truth.}
	\label{fig2}
\end{figure}

The second row of Fig. \ref{fig2} shows the visualization results of the ID 5\_12 area of the ISPRS Potsdam dataset. Compared with the comparative model, our proposed network achieves improved segmentation results in the building area, especially in terms of avoiding background interference. We also test the semantic segmentation performance in this area. The MIoU/AF/OA of the SegNet network is 57.19\%/68.87\%/83.81\%, whereas that of SCAttNet V1 network is 59.50\%/71.16\%/85.04\%. We find that our network achieves higher accuracy in all three metrics compared with the comparative model.

\begin{figure*}[t]
    \vspace{-0.4cm}
	\centering
	\includegraphics[width=6in,height=1.8in]{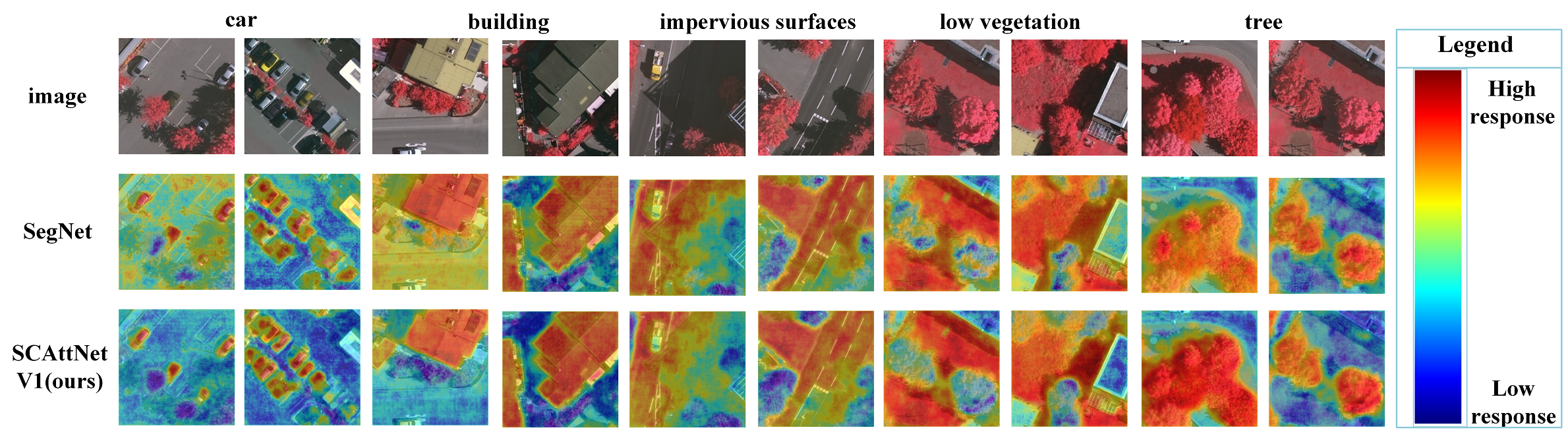}
	\caption{Visualization analysis results of Vaihingen dataset. From left to right: the first two columns represent car categories, the second two columns represent building category, the third two columns represent impervious surfaces category, the fourth  two columns represent low vegetation category, the fifth two columns represent tree category.}
	\label{fig3}
	\vspace{-0.4cm}
\end{figure*}

\subsection{Quantitative analysis of improved interpretability}
To analyze the enhancement of the attention module to the network, we visualize the feature expression of the SegNet network and our SCAttNet V1 network on the ISPRS Vaihingen dataset. We can simply overlay the heatmap of the network before SoftMax operation with an original image, and the relevant areas can be highlighted for a specific category. The visualization results are shown in Fig. \ref{fig3}. After incorporating the attention module, our network can focus on the areas of target and suppress the influence of other categories better than before. As shown in the first  column of Fig.\ref{fig3}, after incorporating the attention module, our network only has a strong response in the car category and other areas are cold toned, which means that the relevance in other areas is very low. However, in the original SegNet network, the impervious area also shows a certain response that may cause interference for the car category. In addition, as shown in the building category in the third column of Fig.\ref{fig3} , the impervious surface area is suppressed after incorporating the attention module. Thus, the impervious surface area avoids classifying into buildings.

\section{Conclusion}
  In this study, we propose a new semantic segmentation network that can adaptively refine features based on the attention module. Experiments on the ISPRS Vaihingen and Potsdam datasets demonstrate the effectiveness of our method. However, some shortcomings remain. The study only uses two common attention modules. Thus, how to design the attention module effectively and capture more discriminative features for semantic inference remains a promising direction for future work.
  
\section*{Acknowledgment}
This work was supported by the National Natural Science Foundation of China [grant numbers 41871276, 41861048, 41871364 and 41871302]. This work was carried out in part using computing resources at the High Performance Computing Platform of Central South University.

\bibliographystyle{IEEEtran}
\bibliography{bare_jrnl}

\end{document}